# Unsupervised Regenerative Learning of Hierarchical Features in Spiking Deep Networks for Object Recognition

Priyadarshini Panda*, and Kaushik Roy
School of Electrical and Computer Engineering, Purdue University
pandap@purdue.edu*

*Abstract*—We present a spike-based unsupervised regenerative learning scheme to train Spiking Deep Networks (SpikeCNN) for object recognition problems using biologically realistic leaky integrate-and-fire neurons. The training methodology is based on the Auto-Encoder learning model wherein the hierarchical network is trained layer wise using the encoder-decoder principle. Regenerative learning uses spike-timing information and inherent latencies to update the weights and learn representative levels for each convolutional layer in an unsupervised manner. The features learnt from the final layer in the hierarchy are then fed to an output layer. The output layer is trained with supervision by showing a fraction of the labeled training dataset and performs the overall classification of the input. Our proposed methodology yields 0.92%/29.84% classification error on MNIST/CIFAR10 datasets which is comparable with state-of-the-art results. The proposed methodology also introduces sparsity in the hierarchical feature representations on account of event-based coding resulting in computationally efficient learning.

## I. INTRODUCTION

"Can machines think?", the question brought up by Turing in his paper, [1], has led to the development of the field of brain-inspired machine learning wherein researchers have put substantial effort in building smarter, more aware devices and technology that have the potential of having human-like understanding. In fact, large scale deep neural network architectures, such as Convolutional Neural Nets (CNNs), have demonstrated unprecedented performance (in terms of classification and recognition accuracy) in a wide range of computer vision and related applications. Such deep networks inspired by the cortical visual processing systems have seen increasing success in recent years due to the availability of more powerful computing hardware (GPU accelerators) and massive datasets for training. Regardless of their success, the substantial computational cost of training and testing such large-scale networks has limited their implementation to clouds and servers. In order to build devices with cognitive abilities, there is a need for specialized hardware with new computational theories. Spiking Neural Networks (SNNs) are a prime candidate for enabling such on-chip intelligence.

Driven by brain-like asynchronous event based computations, SNNs focus their computational effort on currently active parts of the network, effectively saving power on the remaining part, thereby achieving orders of magnitude lesser power consumption in comparison to their Artificial Neural Network (ANN) counterparts [2, 3]. In 2014, IBM research demonstrated a large-scale (>1Million neurons & 256 Million synapses) digital CMOS neurosynaptic chip, TrueNorth [4], which implements a network of integrate-and-fire spiking neurons. However, TrueNorth does not incorporate any information pertaining to the learning mechanisms, which is at present a major constraint for realizing SNNs for real-world practical applications like visual and speech recognition among others. Thus, there is a need to develop efficient learning algorithms that might take the advantage of the specific features of SNNs (event-driven, low power, on-chip learning) while keeping the properties (general-purpose, scalable to larger problems with higher accuracy) of conventional ANN models.

Recent efforts on training of deep spiking networks do not use spike-based learning rules, but instead start from a conventional ANN fully trained using labeled training data, followed by a conversion of the same into a model consisting of simple spiking neurons [5-7]. However, in order to extend the applicability of learning methods, the use of unlabeled data for machine learning is imperative. In the non-spiking domain, unsupervised learning of hierarchical regenerative models such as Auto-Encoders [8] have been successfully used to learn high-level features. The learnt features are then used as inputs to a supervised classification task [9] or to initialize a CNN [10] to avoid local minima. In this paper we develop upon Auto-Encoders where we build a spiking deep CNN by training each layer in the hierarchy in a purely unsupervised manner using the regenerative model and the temporal spike information to update the weights.

SNNs are equipped with unsupervised weight-modification rules like Spike Timing Dependent Plasticity (STDP) that learn the structure of input examples without using labels [11]. However, the network structure that has been successful in achieving competitive classification accuracy for pattern recognition problems is a single-layer SNN, which does not scale well to realistic sized high dimensional images in terms of computational complexity [12]. Moreover, STDP does not support learning hierarchical models that simultaneously represent multiple levels like edges or object parts in a visual context that is fundamental to a deep learning model. We propose regenerative learning using spike-timing information to implement layer-wise weight modification that learn representative levels for each convolutional layer. The features from the final layer of the deep convolutional spiking network (SpikeCNN) are then used for classification tasks.

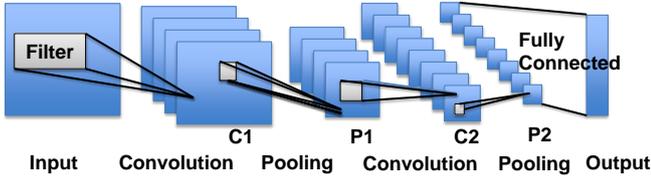

Fig. 1. Standard architecture of a Deep Learning Convolutional Network

## II. PRELIMINARIES

### A. Convolutional Neural Networks

CNNs have proven to be very successful frameworks for image classification tasks [13-15]. Fig. 1 shows the basic CNN structure. They are mainly composed of three main blocks: convolutional layer, spatial sampling/pooling layer and a fully connected layer. The weights of a CNN are convolution kernels. A convolution layer convolves a portion of the previous layer with a set of weight kernels to obtain an array of output maps. The output maps are given by

$$x^k = f(\Sigma_l W^k * x^l) \qquad (1)$$

where $f$ is the neuron's activation function, $x^k$ denotes the activation value of the neurons in the output maps $k$ ($k=1,2...n$), $x^l$ denotes the activation of the neurons in a previous layer's map $l$, $W^k$ are the set of weight kernels and $*$ denotes a 2D- valid convolution operation. The weight kernels are replicated and moved portion-wise over the whole input map. This sharing of weights significantly reduces the number of parameters to be learnt during the training process and thus enables the CNN to be scalable to high-dimensional images. The CNNs are trained in a supervised fashion (showing the training labels) by using standard backpropagation to train the convolutional weight kernels along with the fully connected weights for the final output layer as described in [16].

### B. Unsupervised learning with Auto-Encoders

Unsupervised learning methods are generally used to extract useful features from unlabeled data, to detect oversimplified input representations and remove input redundancies and finally to obtain only robust and discriminative representations of the data. In fact, deep neural network architectures have been built by stacking layers trained in an unsupervised way. This is done to avoid local minima and to increase the network performance [17-19].

Auto-Encoder (AE) methods are one of the most widely used unsupervised feature extractor models for ANNs (non-spiking). AE models are based on the encoder-decoder principle [9]. The input is first transformed into a lower dimensional space (encoder), and then expanded to reproduce the original input (decoder) as shown in Fig. 2. This captures the non-linear dependencies in the input. Each layer is trained on the above principle by feeding the activation from one layer to the next. The basic mathematical formalisms for an AE based training of an ANN are as follows [20]:

$$h = f_\theta = \sigma(Wx) \qquad (2)$$
$$y = f_{\theta'} = \sigma(W'x) \qquad (3)$$
$$W' = W^T \qquad (4)$$

For a given input $x$, the AE first obtains the hidden representation, $h$ using the neuron activation function denoted as σ. This activation value, $h$, is then reverse mapped to reconstruct the input. The weights used in the reverse mapping is generally the transposed form of the weights connecting the input and the hidden layer. Thus, the method uses the same weights for encoding the input and decoding the hidden values, thereby reducing the number of parameters to be learned in the training process. The parameters are then optimized to minimize the error for each training pattern $x_i$ and its reconstruction $y_i$. Note, the training labels are nowhere used in the weight update process. Hence, the method is completely unsupervised. Inspired by the reconstructive model of training, we use the auto-encoder model to update the weights for each convolutional layer by tracking the spike information at the input and modifying the weights such that the original input spike pattern is reproduced. This enables the network to learn hierarchical representative features of the input data.

## III. DEEP SPIKING CONVOLUTIONAL NETWORK: LEARNING AND IMPLEMENTATION

### A. Spiking neuron model

Unlike conventional ANNs where a vector is given at the input layer once and the corresponding output is produced after processing through several layers of the network, SNNs require the input to be encoded as a stream of events. At a particular instant, each event is propagated through the layers of the network while the neurons accumulate the events over time causing the output neuron to fire or spike. Thus, the spike information is used to communicate between the layers of the network. The spiking neuron model used in this work is the Leaky Integrate-and-Fire (LIF) model [31]. The membrane potential $v_{mem}(t)$ of a post synaptic neuron is given by

$$\tau_{RC} \frac{dv_{mem}(t)}{dt} = -v_{mem}(t) + J(t) \qquad (5)$$

where $J(t)$ is the input current and $\tau_{RC}$ is the membrane time constant. The neuron fires when the membrane potential $v_{mem}$ crosses a certain user-defined threshold $v_{th}$. Once a spike is generated, the membrane potential of the neuron is set to the reset potential, $v_{res}$, for a refractory period of $\tau_{ref}$. Once the refractory period is complete, the neuron follows the response shown in equation (5). In our simulations, we discretize the above continuous time equation into 1 ms time steps.

The total synaptic input current received by a neuron is

$$J(t) = \Sigma_i \left( \Sigma_{s \epsilon S_i} w_i \delta(t-s) \right) \qquad (6)$$

where $w_i$ is the synaptic efficacy of the $i^{th}$ synapse, $\delta$ is the delta function that contains the time of arrival of spikes at the $i^{th}$ synapse denoted by $S_i = \{t_i^0, t_i^1 ...\}$.

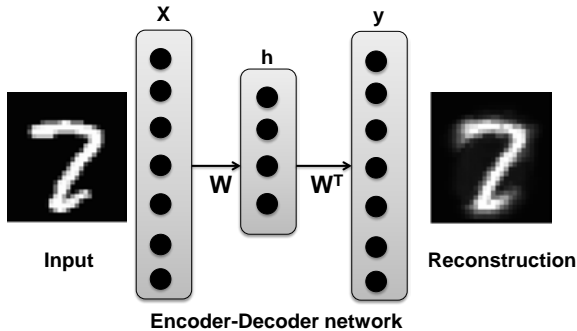

Fig. 2. Auto-Encoder network with input and reconstructed pattern

### B. Error Backpropagation in SNNs

Weight update in Convolutional Deep Learning Networks follow the convolutional backpropagation algorithm which is an extension of the standard stochastic gradient descent rule for feedforward ANNs in the convolutional context [21]. As discussed earlier, in this work, we use the regenerative learning method inspired from AEs to train the hierarchical convolutional layer features of a deep spiking convolutional network (SpikeCNN). Similar to ANNs, in the regenerative learning for deep SNNs, the backpropagation algorithm with gradient descent is employed to update the weights to enable unsupervised layer wise training.

#### B.1. Learning theory

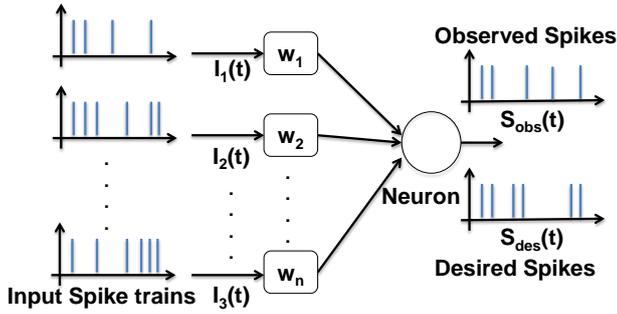

Fig. 3. The learning problem identifies the optimal weight vector so as to achieve the desired spike train from the given input spike pattern.

The learning problem for spiking neurons is illustrated in Fig. 3. There are $n$ input synapses to the spiking neuron, each receiving an independent spike train, and the aim is to determine the n-dimensional synaptic weight vector **w** = [$w_1$ $w_2$ ... $w_n$]$^T$ for the neuron such that it produces the desired spike train $S_{des}(t)$. Let the desired spike train be given by

$$S_{des}(t) = \Sigma_i \delta(t - t_{des}^i) \qquad (7)$$

where $\delta$ is the Dirac delta function and $t_{des}^1, t_{des}^2 ... t_{des}^k$ are the desired spike arrival instants. We now require a learning rule to identify the changes in synaptic weights of the neuron so as to achieve the desired input to output transformation under the constraint that the weight updates be spike induced.

#### B.2. Cost Function

The weight update in ANNs follows the backpropagation algorithm aimed at minimizing a cost function. Similarly, in SNNs we need to define a cost function that would drive the learning process. Let $V(t)$ be the membrane potential of the neuron, with synaptic weight vector **w**, when the given set of input spike trains is fed to it. Correspondingly, let the neuron issue spikes at time instants $t_{obs}^1, t_{obs}^2 ... t_{obs}^{k'}$. Hence, in accordance with (7) the observed spike train can be denoted as

$$S_{obs}(t) = \Sigma_i \delta(t - t_{obs}^i) \qquad (8)$$

Now the error function can be defined as

$$e(t) = S_{des}(t) - S_{obs}(t) \qquad (9)$$

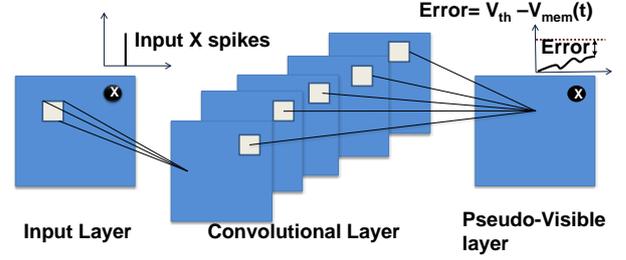

Fig. 4. Layer-wise training of a convolutional layer using Regenerative Learning. If a neuron X in the input layer spikes at a given time instant, the regenerative learning model updates the weights in such a way that the neuron X in the pseudo-visible layer also spikes. This is achieved by propagating the error calculated at the pseudo-visible layer using gradient descent.

However, the desired spike train is not known in an unsupervised learning process. In non-spiking AE models, encoder-decoder method [9] is used to calculate the error that is back propagated through the layers without showing any training labels. Hence, the AE based training is completely unsupervised. Similarly, in the regenerative layer-wise training of SpikeCNN, we add another output layer in addition to the input and the convolutional layer which will be interpreted as a *pseudo-visible layer* that should ideally imitate the input layer patterns as shown in Fig. 4. This will enable us to update weights for the intermediate layers without showing any training labels. In a non-spiking model, the weight update rule is driven by the activation values of the output neurons. In the spiking context, the activation values of an output neuron can be interpreted as its membrane potential. If an input neuron spikes at a given time instant, the corresponding neuron in the output layer should also spike as per the regenerative approach. This can only be achieved if the membrane potential of the output neuron crosses the threshold.

Depending upon the input spike pattern, the error is defined as follows:

- If an input neuron spikes at a given time instant, the error is calculated as the difference in the threshold value ($v_{th}$) and the membrane potential ($v_{mem}$) of the corresponding output neuron in the pseudo-visible layer.

- If the input neuron does not spike, the error is evaluated as the difference in the reset value ($v_{res}$) and the membrane potential ($v_{mem}$) of the corresponding output neuron in the pseudo-visible layer.

Thus, the error function can now be indicated as

$$e(t) = V_{des}(t) - V_{obs}(t) \qquad (10)$$

where $V_{des}$ is $v_{th}$ or $v_{res}$ depending upon the spike event at the input neuron. The cost function corresponding to the synaptic weight vector **w** can now be defined as

$$C(w) = \frac{1}{2} \int_0^T (V_{des}(t) - V(t))^2 dt \qquad (11)$$

where $T$ denotes the duration of the training epoch. The desired weight vector $w_{des}$ is

$$w_{des} = argmin_w C(w) \qquad (12)$$

The learning process thus takes into account the spike information and the inherent latencies. The learning rule follows the gradient descent optimization where the weights are adjusted depending upon the gradient of cost w.r.t the weights. Note that the synapses can be excitatory (leading to increase in membrane potential) or inhibitory (leading to decrease in membrane potential) depending upon the sign of the corresponding synaptic weight as determined by the algorithm.

*B.3. Approximate gradient descent*

The cost function $C(w)$ as given in (11) depends on $V(t)$ which has several discontinuities in the weight space. Hence, as a simplification, we will try to minimize the contribution to the cost function at each time instant independently rather than attempting to minimize the total cost over an entire epoch.

The contribution to the cost-function at time t is obtained by restricting the limits of integral in (11) to an infinitesimally small interval about time *t*. Thus,

$$C(w,t) = \frac{1}{2}\big(V_{des}(t) - V(t)\big)^2 \tag{13}$$

Hence, its gradient with respect to *w* is

$$\nabla_w C(w,t) = -\big(V_{des}(t) - V(t)\big)\nabla_w V(t) \tag{14}$$

Now, the synaptic weight update corresponding to the activity observed at time *t* is given as

$$\Delta w(t) = -\eta \nabla_w C(w,t) = -\eta\big(V_{des}(t) - V(t)\big)\nabla_w V(t) \tag{15}$$

where $\eta$ is the user-defined learning rate. Now, the discontinuities in $V(t)$ due to the LIF neuron model will render $\nabla_w V(t)$ undefined for some particular values of *w*. In [22], the authors have implemented a weight update rule and have used certain approximations to overcome these non-linear dependencies. In this work, we invoke similar approximations that would allow us to replace $\nabla_w V(t)$ with appropriate quantities.

In non-spiking context, the weight update for a synapse connecting neuron *i* and *j* [23] (with exponential activation function) is computed as

$$\Delta w_{ij} = -\eta\, \partial E/\partial w_{ij}; where\ \partial E/\partial w_{ij} = \delta_j o_i \tag{16}$$

$$\delta_j = E\ o_j \qquad\quad if\ j\ is\ an\ output\ neuron$$
$$\quad = (\Sigma_l \delta_l w_{jl})\, o_j \quad if\ j\ is\ a\ hidden\ neuron \tag{17}$$

Here, *o* denotes the activation value of neuron. Comparing (15) and (16), $\nabla_w C(t)$ is equivalent to $\partial E/\partial w_{ij}$. Interpreting the activation value of the neuron as the membrane potential in the spiking context, we can now replace $o_j$ in the above equations with the membrane potential of a neuron ($V(t)$) at a given time instant. Thus, the weight update equations for SNN with approximate gradient descent can be written as

$$\Delta w_{ij}(t) = -\eta\, \delta_j(t) V_i(t) \tag{18}$$

$$\delta_j(t) = (V_{des_j}(t) - V_j(t))\, V_j(t)\ \ if\ j\ is\ output\ neuron$$
$$\quad = (\Sigma_l \delta_l(t) w_{jl}) V_j(t) \qquad if\ j\ is\ hidden\ neuron \tag{19}$$

In summary, the approximate gradient descent assumes the LIF neuron model to be an equivalent standard activation model in non-spiking context. The gradient calculation is then carried out by using the membrane potential of a neuron at a given time instant as the activation value. The inherent error-resiliency of these neural networks allows us to use such approximate models.

*C. Regenerative Learning for Spike-based Convolutional Auto-Encoder*

Convolutional Auto-Encoders (CAEs) differ from conventional AEs as their weights are shared among all locations in the input preserving spatial locality. The CAE architecture with weight sharing is shown in Fig. 4. The reconstruction represented by the pseudo-visible layer is due to a linear combination of basic image patches based on the activations in the convolutional layer.

For a mono-channel input *x*, the convolutional layer representation of the $k^{th}$ feature map is given by

$$h^k = \sigma(x * w^k) \tag{20}$$

Here, $h^k$ denotes the membrane potential of the neuron ($v_{mem}(t)$) in the convolutional layer, $\sigma$ denotes LIF model discussed earlier to calculate $v_{mem}(t)$ at a given time instant *t*, *x* contains the spike information from the input layer neurons and * denotes the convolution operation. When $h^k$ crosses $v_{th}$, the spikes generated at the neurons of the convolutional layers serve as input for the pseudo-visible layer. The reconstruction is obtained using

$$y = \sigma(\Sigma_k \varepsilon(h^k) * \widetilde{w}^k) \tag{21}$$

where $\widetilde{w}^k$ denotes the transpose (or flip in both dimensions) operation, $\varepsilon(h^k)$ denotes the spike information corresponding to the convolutional layer and *y* gives $v_{mem}(t)$ of the neurons in the pseudo-visible layer. The convolution of a *m* x *m* matrix with a *n* x *n* matrix may result in an *(m-n+1)* x *(m-n+1)* matrix (valid convolution) or *(m+n-1)* x *(m+n-1)* (full convolution). In our simulations, we perform a valid convolution from input to the convolutional layer and a full convolution from convolutional to pseudo-visible layer in order to get a map of the same size as the input layer. All spike-based models and calculations for implementing convolutional network remain the same as described in earlier sub-section (III-B). Earlier in (6), the input synaptic current was given by the weighed summation of spike inputs. In this case, the only difference is that the synaptic current is obtained by convolving the spike information of an image patch with the weight kernel.

Please note that the $h^k$ and *y* are calculated at every time instant of a given epoch and the error is calculated from the difference in the $v_{mem}(t)$ of the neurons in the input and pseudo-visible layer. It is clear that the regenerative learning would activate the neurons in the pseudo-visible layer such that they imitate the input layer spike pattern. Thus, the cost function to minimize is the mean squared error (MSE) given by

$$C(w) = \frac{1}{2n}\Sigma_{i=1}^{n}\big(v_{des}(t) - v_{mem}(t)\big)^2 \tag{22}$$

where *n* is the total number of neurons in the input /pseudo-visible layer. As mentioned earlier, we minimize the error at each time instant rather than over the entire epoch duration.

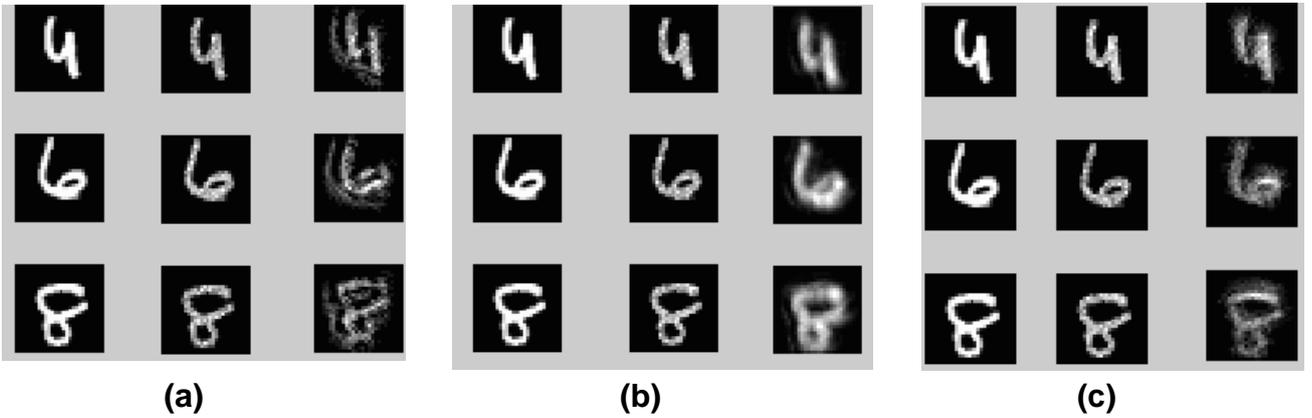

Fig. 5 Reconstructed patterns observed after training the first convolutional layer of MNIST_2C with regenerative learning for different $v_{th}$ and $I_{rate}$ values {Original pixel image (Column 1), Spike input image (Column 2), Reconstructed image (Column 3)} (a) MNIST_2C initialized with $v_{th}$= 1.0, $I_{rate}$ =100 Hz (b) MNIST_2C with parameters ($P_2$) $v_{th}$= 1.2, $I_{rate}$ =100 Hz (c) MNIST_2C with parameters ($P_1$) $v_{th}$= 0.8, $I_{rate}$ =75 Hz

Now the approximate gradient descent model discussed in sub-section *B.3* is used to calculate the weight updates. The gradient of the cost function now involves convolution operations and is given by

$$\frac{\partial C(w)}{\partial W} = x * \delta h^k + \widetilde{\varepsilon(h^k)} * \delta y \qquad (23)$$

$\delta h^k$ and $\delta y$ are the gradients for the convolutional and pseudo-visible layer neurons which are evaluated using the approximate model. $\delta h^k$ is calculated using (19) for hidden neuron and $\delta y$ using (19) for output neuron. Several such convolutional AEs can be assembled together to construct a deep CNN hierarchy. Each convolutional layer in the hierarchy is trained separately with the regenerative method described above.

### D. Average Pooling

For CNNs, a pooling layer [24, 25] is often introduced after the convolutional layers to obtain translational invariance. The average pooling or subsampling layers combine the responses from multiple neurons in the convolutional layer into one. The representation of the averaging layer is identical to (1), except that the kernels consist of uniform weights fixed to *1/size($w^k$)*, where *size($w^k$)* is the size of the sampling window. After training the convolutional layer with the regenerative method, the membrane potentials ($h^k$) of the neurons across all feature maps are obtained corresponding to a training input for a given instant of an epoch. The membrane potentials ($h^k$) within the sampling window are averaged to obtain the output membrane potential of the neuron ($p^k$) in the pooling layer for the given time instant. When $p^k$ crosses $v_{th}$, the spikes generated at the neurons of the pooling layer serve as input for training the next convolutional layer. It is clear that the averaging operation down-samples the convolutional layer representation while conserving the spike information and inherent latencies from the previous layer.

The regenerative auto-encoder based learning, thus, trains several convolutional layers in layer-wise fashion which can be assembled together to form a deep hierarchy [27: SAE]. Each layer receives its input from the previous layer as described above. The assembled convolutional layers can be used to initialize a CNN with the same topology prior to a fully connected stage.

### E. Supervised training with labels for classification

The Fully Connected layer (FC) at the end of the CNN combines the inputs from the feature maps in the previous layer to perform classification of the overall inputs at the output layer. The training of the fully connected layer cannot follow the encoder-decoder principle (regenerative learning) as the aim here is to classify rather than to obtain abstract representations of the input. At this stage, the spike information from all end layer maps are concatenated into a vector which serves as input from the FC layer to the output layer.

The training labels are used to fix the spike pattern of the output layer neurons that will drive the error backpropagation to calculate the weight updates. The weight update rules follow the same equations as described in the approximate gradient descent method in sub-section *B.3*. For a given label, a Poisson spike pattern of a particular frequency is generated that serves as the desired spike train for the corresponding output neuron for all training inputs with that label. The remaining output neurons have no spike events. The errors are calculated as per the spike events at the output layer neurons. This method ensures that the weights connected to the desired output neuron are potentiated while inhibiting the activity of the remaining connections. In our simulations, we fix the frequency of the desired spike train for a training label at 30 Hz. After training is done, an input is presented as a stream of events at the input layer of the SpikeCNN. At the end of the time duration, the spike responses from the output layer neurons are monitored. The output neuron with the highest response is the class predicted by the network for the given input.

## IV. EXPERIMENTAL RESULTS

We evaluate our proposed learning on two datasets: MNIST [16] and CIFAR10 [26]. MNIST is a standard dataset of handwritten digits that contains 60,000 training and 10,000 test patterns of 28 x 28 pixel sized greyscale images of the digits 0-9. CIFAR10 is a more challenging dataset that consists of 60,000 colored images belonging to 10 classes. Each image has 32x32 pixels. We used the first 50,000 images for training and last 10,000 images for testing. Regenerative learning is used to train spike-based convolutional AEs which are then used to initialize a SpikeCNN with the same topology. The final fully connected layer of the SpikeCNN is then trained with a fraction of training labels from the entire dataset to perform classification. The input is presented as Poisson distributed spike train with firing rates proportional to the intensity of pixels [5]. In the experiments, the epoch duration or the time for which an input image is shown to the network is 250ms. During layer-wise training, each input is presented multiple times (depending upon the depth of SpikeCNN) and the weights are updated each time to minimize the reconstruction error. Since each layer receives its input from the previous trained layer, this process helps in maintaining an adequate firing rate for each layer. This ensures spike propagation as we go deeper into the network. The membrane potentials of all neurons in the convolutional and pseudo-visible layers are all set to $v_{res}$ before presenting a new training pattern.

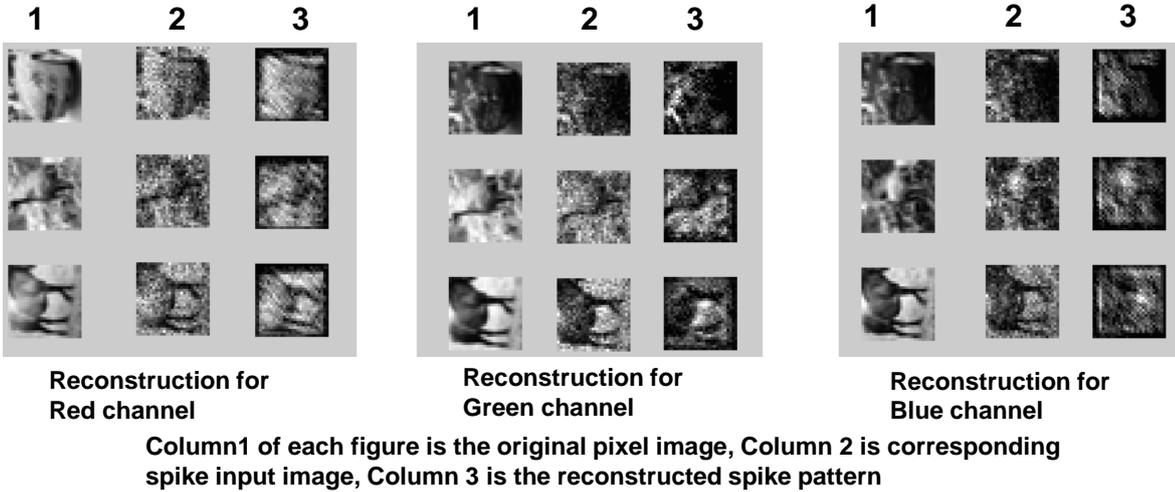

Fig. 6 Reconstructed pattern observed after training the first layer of CIFAR_3C initialized with $P_2$. CIFAR_3C has 3 maps at the input layer corresponding to the 3 color channels (Red, Green, Blue). The figures show the reconstruction of the input pattern at the pseudo-visible layer corresponding to the 3 separate channels.

*A. Network Architecture and Parameters*

We implemented a SpikeCNN for MNIST (MNIST_2C) with 2 convolutional layers: 28x28-12c5-2a-64c5-2a-10o. The input layer is 28x28. Both convolutional layers use 5x5 kernel size with 12 and 64 maps, respectively. A 2x2 average pooling window is used after each convolutional layer. The final features from the second averaging layer are then fully connected to a 10-neuron output layer. SpikeCNN for CIFAR10 (CIFAR_3C) consists of 3 convolutional layers: 32x32x3-32c5-2a-32c5-2a-64c4-10o. In this case, the input layer has 3 maps corresponding to the 3 color channels RGB. The first and second convolutional layer have 5x5 sized kernels with 32 maps while the third layer has 4x4 kernel with 64 maps. The features from the third layer are directly fed to the output layer without any average pooling. Please note that we do not use any data augmentation or normalization techniques like dropout [27] in the SpikeCNN implementation.

The network parameters like input rates ($I_{rate}$) and threshold values ($v_{th}$) for the networks were set by trial and error by cross-validating a few times to get the lowest reconstruction error from the input image layer. Fig. 5 shows the reconstructed image patterns formed at the pseudo-visible layer after training the first convolution layer of MNIST_2C with regenerative learning for different values of $v_{th}$ and $I_{rate}$. Visually, we can inspect that Fig. 5(b) with parameters $v_{th}=$ 1.2, $I_{rate}$ =100 Hz ($P_2$) and Fig. 5(c) with $v_{th}=$ 0.8, $I_{rate}$ =75 Hz ($P_1$) give more convincing reconstruction than Fig. 5(a) $v_{th}=$ 1.0, $I_{rate}$ =100 Hz. We use the parameters $P_2$, $P_1$ corresponding to Fig. 5 (b), (c) to initialize MNIST_2C and evaluate the classification accuracy in both cases. Similarly, Fig. 6 shows the reconstructed image patterns from the first convolutional layer of CIFAR_3C for the three color channels separately. The parameters used are $v_{th}=$ 1.2, $I_{rate}$ =100Hz to initialize CIFAR_3C and obtain the classification accuracy. The figures show the accumulated spike count over 250ms of simulated time. Please note that though the deeper layers in MNIST_2C/CIFAR_3C have significantly larger number of maps than that of initial layers in both the configurations, the training time still remains same due to the down-sampling of input size with average pooling.

*B. Reconstruction error across network layers*

For quantitative evaluation, we plot the reconstruction errors obtained at each layer of the network (MNIST_2C, CIFAR_3C) as shown in Fig. 7 for the network parameters discussed above. In order to ensure propagation of spike information across layers, we present the input data 3/5 times while training every convolutional layer of MNIST_2C/CIFAR_3C respectively. The reconstruction error plotted in Fig. 7 is the aggregate loss over the multiple presentation of the training data. We use the squared Euclidean distance of the difference in the spike events as a measure of loss. It is clearly seen that error observed for MNIST_2C with parameters $P_1$ is higher than that of $P_2$ which supports the visual reconstruction patterns shown in Fig 5(b), (c). A noteworthy observation here is that the reconstruction error in both networks decreases as we move towards deeper layers. In [28], the authors have shown that in the context of deep ANNs, blurring an image enables better reconstruction as the network then learns more general representations. Learning the finer details may lead to overfitting increasing the reconstruction error. As we apply convolution and average pooling in the initial stage, certain information from the original image is lost in this process. Thus, we can interpret that the deeper layers work on slightly blurred details of the original image causing lower reconstruction error.

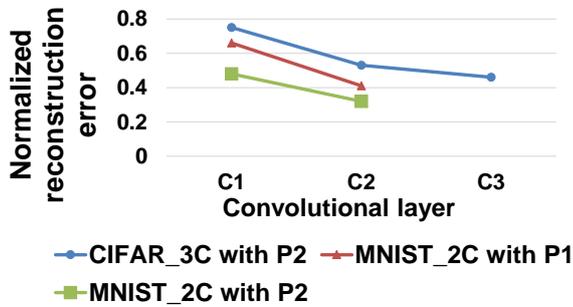

Fig. 7 Reconstruction errors at different layers of the SpikeCNN architecture

C. Classification Accuracy

After training the convolutional layers by optimizing the reconstruction error for the entire training dataset, the features from the final layer are then fed to the output layer. The weights at the output are trained in a supervised manner by showing training labels as discussed in Section III (*E*). However, in the supervised case, we use only a subset of the training data to train the final layer. During testing, an input test pattern is presented two times to the trained SpikeCNN. The spikes at the output neurons are aggregated over two simulations and the neuron with the highest response is the predicted class for the given test input. Since the pixel intensity values are converted to Poisson spike trains, the accuracy can differ for different spike timings. Thus, the accuracies are averaged over five iterations of presentation of the entire testing dataset. Fig. 8 shows the classification error for MNIST_2C initialized with parameters $P_1$, $P_2$ and CIFAR_3C with $P_2$ as the size of the labelled training subset is varied. For MNIST_2C with $P_2$, the lowest error achieved by showing all the 60000 training labels at the final layer is 0.92% (99.08% classification accuracy). It is clearly seen that the error decreases significantly as the size of the training set is increased from 500 to 20000. However, for subsets > 20000, the error almost remains the same. For MNIST_2C with $P_1$/ CIFAR_3C with $P_2$, the minimum error obtained showing 20000 labels at the final layer is 1.81%/29.84%. In [29]/[30], the authors have implemented a spiking deep network by converting a deep static CNN to SNN and have achieved 22.57%/0.86% error on CIFAR-10/MNIST dataset. The fact that our network performs favorably incorporating the inherent latencies of a spiking neuron model in the learning process suggests that the regenerative learning scheme can be used to train deep spiking networks to obtain state-of-the-art results.

D. Sparsity with Regenerative Learning

The spike-based regenerative learning scheme, on account of event-based coding, introduces *sparsity* over the convolutional layer feature representations. Since the learning is based on spiking activity of the neurons, the output at the pseudo-visible layer is reconstructed using only the maximally active neurons in the feature maps of the convolutional layers. As a result, the reconstruction error at each time instant of the training epoch is back-propagated through these active neurons. Basically, the sparse event based computation acts as a regularizer that prevents learning of over-complete representations of the input. In other words, the sparsity in features decreases the number of filters or weight kernels required to reconstruct the input thereby forcing the filters to be more general. Fig. 9 shows the feature maps learnt for MNIST_2C (with $P_2$) with accumulated spike counts over 250 ms for a particular input pattern. It is evident that the feature maps in both convolutional layers have sparse active units. Only a smaller section of the SpikeCNN is active in a training epoch. Thus, we can effectively save power on the remaining idle or inactive portions of the network. Sparsity in learning has a key role in reducing the overall power consumption by decreasing the spike rate in SNN architectures which is one of the main reasons to use SNN over ANN. Fig. 9 also shows the weight kernels learnt at the first layer from the input. Visually, we can interpret that the weights are more diverse and global.

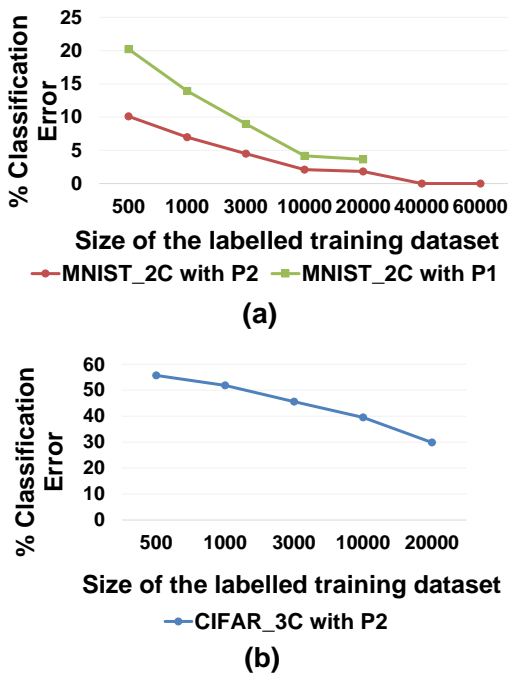

Fig. 8 Classification error as the size of the labelled dataset for the supervised training of output layer is varied

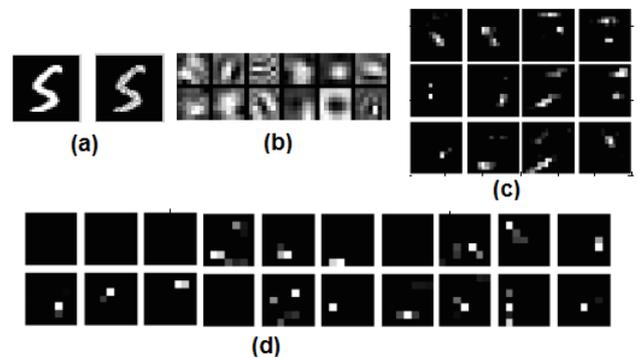

Fig. 9 (a) MNIST training input to MNIST_2C initialized with $P_2$: Original pixel image (left), Spike input image (right) (b) The weight kernels learnt at the input layer (c) 12 feature maps showing the sparse representations of maximally active spiking neurons in the first convolutional layer of MNIST_2C (d) 20 of 64 feature maps in the second convolutional layer of MNIST_2C with sparsely active neurons

## V. CONCLUSION

We introduced a spike-based learning scheme to train Spiking Deep Networks (SpikeCNN) for object recognition problems using leaky integrate-and-fire (LIF) neurons. The regenerative model learns the hierarchical feature maps layer-by-layer in a deep convolutional network in an unsupervised manner. Once the convolutional layers are learnt, the features are then fed to an output layer trained in a supervised manner by showing a fraction of the labeled training dataset. The output layer performs the overall classification of the input. While previous work on deep spiking networks have examined the conversion of ANNs to SNNs, we build a spiking deep CNN from scratch with the proposed learning using spike-timing information and inherent latencies to implement layer-wise weight modification. Our experiments on the MNIST and CIFAR-10 dataset demonstrate comparable classification accuracy with state-of-the-art results. Also, the sparsity in representations introduced with regenerative learning suggests overall power savings in the learning process which generally takes up a significant part of the time the network is used. Finally, the SpikeCNN system developed with our proposed learning with the current initialized parameters is in its first generation, and we expect its accuracy on CIFAR-10 to improve as we gain experience with the method and tune the network for better parameters.